\documentclass[letterpaper, 10 pt, conference]{ieeeconf}  

\IEEEoverridecommandlockouts                              

\usepackage{graphics} 
\usepackage{subfiles}
\usepackage{csquotes}
\usepackage{gensymb}
\usepackage{pifont}
\usepackage{textcomp}
\usepackage[table]{xcolor}
\usepackage[english]{babel}
\usepackage{biblatex}
\usepackage{pdfpages}
\usepackage{blindtext}
\usepackage{soul}
\usepackage{hyperref}
\usepackage{graphicx}
\usepackage{mathtools}
\usepackage{multirow}

\graphicspath{{images/}{../images/}}
\usepackage{subcaption}
\usepackage[font={footnotesize}]{caption}
\usepackage[export]{adjustbox}

\usepackage{amsmath}
\usepackage{amssymb} 
\usepackage{amsfonts}
\usepackage{bm} 
\usepackage{siunitx}
\usepackage{cleveref}

\title{\Large \bf
An RLS-Based Instantaneous Velocity Estimator for Extended Radar Tracking 
}

\author{Nikhil Bharadwaj Gosala$^1$ and Xiaoli Meng$^2$ \\ $^1$ETH Z{\"u}rich, Switzerland \quad $^2$Hyundai-Aptiv Autonomous Driving JV, Singapore 
        \\[-.5ex]
}

\begin{document}

\maketitle
\thispagestyle{empty}
\pagestyle{empty}

\vspace{-2pt}

\begin{abstract}
Radar sensors have become an important part of the perception sensor suite due to their long range and their ability to work in adverse weather conditions. However, several shortcomings such as large amounts of noise and extreme sparsity of the point cloud result in them not being used to their full potential. In this paper, we present a novel Recursive Least Squares (RLS) based approach to estimate the instantaneous velocity of dynamic objects in real-time that is capable of handling large amounts of noise in the input data stream. We also present an end-to-end pipeline to track extended objects in real-time that uses the computed velocity estimates for data association and track initialisation. The approaches are evaluated using several real-world inspired driving scenarios that test the limits of these algorithms. It is also experimentally proven that our approaches run in real-time with frame execution time not exceeding \SI{30}{\milli\s} even in dense traffic scenarios, thus allowing for their direct implementation on autonomous vehicles.
\end{abstract}

\section{Introduction}
\label{sec:introduction}
Autonomous vehicles employ a multitude of sensors to effectively perceive the environment around them. The radar has become an essential part of this myriad of sensors due to its ability to precisely estimate the position and velocity of various targets up to a very long range, and to work reliably in adverse weather conditions like dense fog and heavy rain which trouble other sensor modalities.
At the same time, radar has its own set of shortcomings such as the presence of a large amount of noise from sources such as wheel wells, interference from other radar systems, and the extreme sparsity of the point cloud due to its relatively large angular resolution~\cite{wei-radar, nvidia-radar, dickmann-radar, yurtsever-radar}.

Traditionally, point trackers have been used to estimate the position and velocity of objects, and to track them over multiple frames in the radar point cloud.
However, owing to the development of high-resolution radars that enable sensors to return multiple detections per target, the use of point trackers results in the generation of incorrect and phantom tracks especially when dealing with long objects where the point object assumption is often violated~\cite{eotsummary}. Additionally, classical filter-based approaches that assume only one return per object are inapplicable to this new problem domain~\cite{Scheel.2018}.
To address this issue with point trackers, the traditional point object model has been replaced by the extended object model wherein all points belonging to an object are processed together to determine its position, velocity and extents~\cite{eotsummary}. 
Extended Object Tracking (EOT) thus refers to the recursive estimation of the position, velocity, and extents of objects that provide multiple noisy returns per frame, and whose shape is unknown and can vary over time~\cite{eotsummary}.

The extended object problem has usually been solved either by using pre-processing techniques that represent all points from an object using a single meta-point, or by designing models that explicitly account for multiple returns from target objects~\cite{Scheel.2018}.
The former set of solutions mainly extract the shape by clustering~\cite{7161279, Henriksson2016RadarBT, 7535453} and by shape fitting~\cite{7518649, 7918865}, whereas the latter approaches work by modelling the shape using spatial distributions \cite{1512732, 6641184}, by adopting physics-based modelling \cite{Scheel2016MultisensorMT, dirscattering, 6237574}, and by modelling target objects as a set of reflection centres \cite{6237597, 4290130}. 
Although the latter set of approaches do not discard useful information by representing multiple points by a single meta-point and are more accurate than the former, they usually are very restrictive in their application, require a lot of human modelling and engineering, and do not run in real-time. Additionally, the former set of solutions are often not perfect and excessively rely on a fool-proof clustering algorithm, use a hard-association algorithm which limits the association of one cluster to only one track, and use insufficient metrics to associate clusters to tracks.


In this paper we present an end-to-end pipeline to perform EOT that runs in real-time, does not extensively rely on fool-proof clustering, and uses both position and instantaneous velocity to perform the one-to-many track-cluster association. 
Furthermore, as one of the key contributions of this paper, we also propose a novel Recursive Least Squares (RLS)-based algorithm to estimate the instantaneous velocity of a cluster in real-time that allows for accurate track-cluster association, removes the need to perform computationally expensive non-linear state updates, and allows for the estimation of the true velocity even in frames with a large amount of clutter.

The paper is organised as follows: Section~\ref{sec:eot} describes our end-to-end EOT pipeline in detail, Section~\ref{sec:rls} presents our novel instantaneous velocity estimation algorithm, Section~\ref{sec:experiments} evaluates the performance of our velocity estimation algorithm and our EOT pipeline using real-world scenarios, and the paper is concluded in Section~\ref{sec:conclusion}. A video showcasing the real-world performance of our RLS-based instantaneous velocity estimation algorithm can be found \href{https://www.youtube.com/watch?v=B5SPXXkZpz8}{online}\footnote{\url{https://www.youtube.com/watch?v=B5SPXXkZpz8}}.
\section{Extended Object Tracking Pipeline}
\label{sec:eot}

In this section we present our real-time end-to-end pipeline to track dynamic extended objects using data from the radar alone, the measurements of which are composed of the position, the Doppler velocity, and the bearing of the reflection point w.r.t. the sensor.

\begin{figure}
	\centering
    \includegraphics[width=\columnwidth]{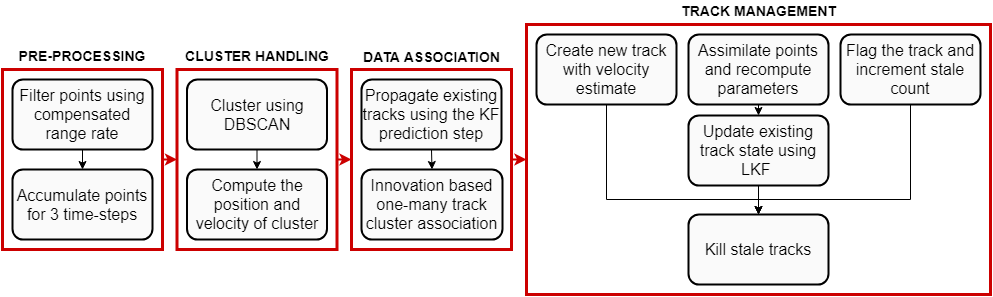}
    \setlength{\abovecaptionskip}{-5pt}
    \setlength{\belowcaptionskip}{-20pt}
  \caption{The overall extended object tracking pipeline. Parts of the pipeline have been grouped together for ease of understanding. Image best viewed on a computer.}
  \label{fig:overall-pipeline}
\end{figure}

\subsection{Overall Pipeline}
\label{subsec:overall-pipeline}
An EOT pipeline enables the estimation of the length and width of tracked objects in addition to parameters like position and velocity which allows for the precise representation of the world around an autonomous vehicle, enabling accurate path planning and robust collision avoidance. The tracks generated by this tracker are used in conjunction with those from other sensor modalities to generate very precise final tracks. However, when all the other sensors malfunction, the tracks from radar can be used either to continue operation at a reduced speed, or to allow for the safe stoppage of the vehicle. Our EOT pipeline, depicted in Figure~\ref{fig:overall-pipeline}, can broadly be classified into four categories namely, \textit{Pre-processing}, \textit{Cluster Handling}, \textit{Data Association}, and \textit{Track Management}, each of which is detailed below.

\subsection{Pre-processing}
\label{subsec:preprocessing}
The pre-processing phase encompasses two phases namely, \textit{Static Point Removal} and \textit{Frame Accumulation}. In the former, only dynamic points, i.e., points having a compensated range rate of at least \SI{0.5}{\meter/\s} are retained which facilitates the removal of noise and other indiscernible points from the input point cloud. However, this filtration also removes non-noise points with a range rate less than this threshold, but it has experimentally been noted by us that their omission does not affect the performance of the tracker. 
Furthermore, we increase the density of the radar point cloud by accumulating it over $3$ time steps because the extreme sparsity of the point cloud makes detection and tracking of objects from only one time step very challenging. Doing so not only makes detection and tracking easier but also provides a good prior for estimating the instantaneous velocity of clusters, as will be discussed in section~\ref{sec:rls}. However, accumulation introduces significant distortion in an object which can skew its position and extents estimates. For instance, when the radar refreshes at \SI{14}{\hertz} and the object's speed is \SI{14}{\meter/\s}, the distortion over $3$ time steps is as large as \SI{3}{\meter}. This distortion is corrected using its velocity estimate computed in section~\ref{subsubsec:velocity} as shown in equation~\eqref{eqn:motion-distortion-comp}, and this correction is performed after the clustering phase, detailed in section~\ref{subsec:cluster-handling}, is completed.
\begin{gather}
\label{eqn:motion-distortion-comp}
    {}^{t_1}\bm{T}_{n}^{t_2} = -\bm{v}_{n}(t_2 - t_1)
\end{gather}
Here, ${}^{t_1}\bm{T}_{n}^{t_2}$ is the transformation that corrects the distortion between time steps $t_1$ and $t_2$ in the n\textsuperscript{th} cluster, $\bm{v}_{n}$ is its velocity as computed in section~\ref{subsubsec:velocity}.

\subsection{Cluster Handling}
\label{subsec:cluster-handling}
\begin{figure}
  \centering
  \begin{subfigure}[b]{0.30\columnwidth}
  \centering
  	\includegraphics[width=\linewidth]{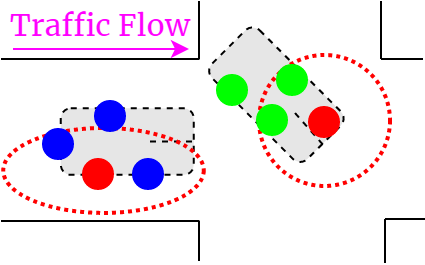}  
    \caption{}
    \label{fig:clustering-traffic-flow}
  \end{subfigure}
  \hspace{1em}
  \begin{subfigure}[b]{0.63\columnwidth}
  \centering
  	\includegraphics[width=\linewidth]{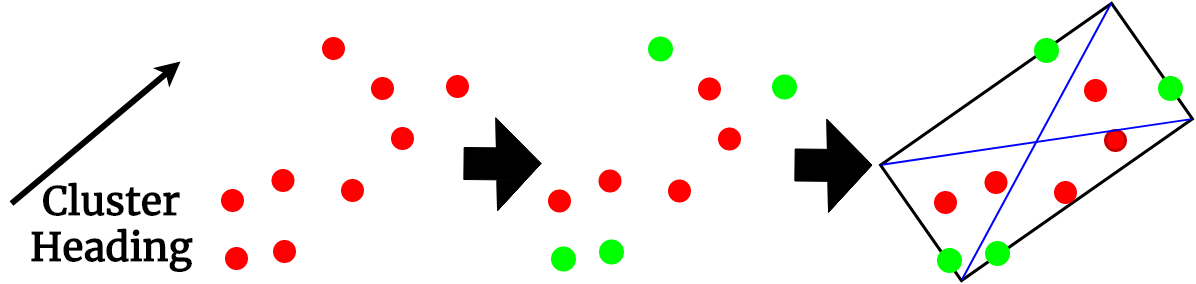}
    \caption{}
    \label{fig:bounding-box}
  \end{subfigure}
  \setlength{\abovecaptionskip}{-10pt}
  \setlength{\belowcaptionskip}{-20pt}
  \caption{(a) The elliptical and circular grouping when the point, shown in red, is in an area with a defined and ambiguous traffic flow direction respectively. (b) Steps involved in estimating the centre of a cluster. The extreme points (green) are computed using the cluster, after which a rectangular bounding box is drawn through these points. The centre is then the geometric centre of this bounding box. The arrow shows the heading of the cluster.}
\end{figure}

In the second phase of the pipeline, all points from the accumulated frames are clustered using the DBSCAN algorithm~\cite{ref:dbscan}, and the position and velocity metrics are computed for each cluster. DBSCAN is a density-based clustering algorithm that groups points that are densely packed together and marks points lying in low density regions as outliers. DBSCAN is preferred to other clustering algorithms because of its adaptability to varying cluster shapes, and its efficient handling of outliers and clutter in the input dataset. To allow for an accurate grouping of vehicles on the road wherein most of them have their lengths aligned to the flow of traffic, we use the available pre-annotated high-definition (HD) map to group points within an elliptical region whose major axis is aligned with the flow of traffic instead of the traditional circular grouping of points. When the flow of traffic is unclear, e.g. in intersections, we revert to using the traditional circular region to group points as shown in Figure~\ref{fig:clustering-traffic-flow}. Lastly, as detailed later in section~\ref{subsec:data-assoc}, because we employ a many-one data association algorithm to associate clusters to existing tracks, it is acceptable for the clustering phase to not be accurate and not cluster all returns from an object together. Such an approach also ensures accurate tracking even when the vehicle is not aligned to the traffic flow direction, e.g. when the vehicle is changing lanes or is pulling out onto the road. 

Once the grouping is complete, the position and velocity of the clusters are estimated in the world frame and the procedure to compute them is detailed below.

\subsubsection{Position}
\label{subsubsec:position}
All points in the cluster are represented by a meta-point that is assumed to be in the geometric centre of the cluster. The geometric centre of the cluster is obtained by computing the centre of the tightest possible rectangular bounding box around all points in the cluster as shown in Figure~\ref{fig:bounding-box}. The geometric centre is chosen over the centroid because the distribution of points over the target object is generally not uniform resulting in distorted centroid estimates. The position innovation ($I_{\text{pos}}$) is then the squared distance between the predicted position of the track and the geometric centre of the cluster. Mathematically, 
\begin{equation}
\label{eqn:inno-pos}
    I_{\text{pos}} = (x_{\text{track}} - x_{\text{cluster}})^2 + (y_{\text{track}} - y_{\text{cluster}})^2
\end{equation}
where $x_{\text{track}}$ and $y_{\text{track}}$ denote the estimated position of the centre of the track, and $x_{\text{cluster}}$ and $y_{\text{cluster}}$ denote the coordinates of the centre of the cluster.

\subsubsection{Velocity}
\label{subsubsec:velocity}
Unlike position which is directly obtained from the radar sensor, the velocity of the cluster needs to be explicitly computed using the Doppler velocity and the bearing w.r.t. the sensor of each point in the cluster. We use a novel RLS-based algorithm, presented in detail in section~\ref{sec:rls}, to compute the instantaneous axis-aligned velocity of the cluster. The velocity innovation ($I_{\text{vel}}$) is then the squared difference between the velocity of the track and the velocity of the cluster. Mathematically,
\begin{equation}
\label{eqn:inno-vel}
    I_{\text{vel}} = (v_{x}^{\text{track}} - v_{x}^{\text{cluster}})^2 + (v_{y}^{\text{track}} - v_{y}^{\text{cluster}})^2
\end{equation}
where $v_{x}^{\text{track}}$ and $v_{y}^{\text{track}}$, and $v_{x}^{\text{cluster}}$ and $v_y^{\text{cluster}}$ represent the axis-aligned velocity of the track and cluster respectively.

\subsection{Data Association}
\label{subsec:data-assoc}
Once the clustering phase is complete, the next part of the pipeline is to ascertain whether any of the newly computed clusters are point returns from objects that are already being tracked by the tracker. This phase is called \textit{Track-Cluster Association} and is not trivial because we do not know which data points come from which object and vice-versa. We use two parameters - position and velocity - to compute the association between the tracks and the clusters. A cluster is associated to a track if the innovation between the parameters of the track predicted using the constant velocity (CV) motion model and those of the cluster is less than \SI{9}{\meter^2} and \SI{9}{\meter^2/\second^2} respectively, and the sum of innovations of the two parameters is the least amongst all potential tracks. It is worth noting that because we use a one-many data association algorithm, multiple clusters can be associated to the same track which improves the performance in case of long objects and reduces the reliance on the need for an accurate clustering algorithm, as depicted in Figure~\ref{fig:one-many-association}.

\subsection{Track Management}
\label{subsec:track-management}
The track-cluster associations generated from the previous phase are used to either create, update, or delete tracks based on their association history. This section explains how the track state is maintained and how the track-cluster association is handled.

\subsubsection{Track State}
\label{subsec:track-state}
The state of a track is defined using a $6$-dimensional state vector that keeps track of its position, velocity, and extents. Mathematically, it is represented as $\{x, y, v_{x}, v_{y}, l, w\}$, where $x$ and $y$ denote the position of the centre of the track in the XY plane, $v_{x}$ and $v_{y}$ denote the axis-aligned velocity of the track, and $l$ and $w$ denote the length and width of the track respectively. The track is maintained using a Linear Kalman Filter (LKF) with the propagation step using a CV motion model to propagate the tracks. An LKF is used because all the state parameters can be updated in a linear fashion directly from the raw or derived measurements and is much more computationally efficient as compared to its non-linear counterparts. Our LKF uses the standard update equations and they have been omitted in the interest of space.

\subsubsection{Cluster not associated to any track}
\label{subsubsec:cluster-no-track}
When a cluster is not associated to any track, a new track is created with the state parameters initialised using the corresponding values of the cluster. Since the majority of the moving objects on roads are vehicles, we initialise the $l$ and $w$ to \SI{4}{\meter} and \SI{2}{\meter} respectively. To prevent noise and other random reflections from interfering with the track list, the new track is initially set to an \textit{invalid} state which becomes valid when the track is associated to at least one cluster for $3$ iterations.

\subsubsection{Cluster associated to a track}
\label{subsubsec:cluster-track}
A cluster is associated to a track when the innovation of both position and velocity is less than \SI{9}{\meter^2} and \SI{9}{\meter^2/\second^2} respectively, and the sum of their innovations is the least for that track. Because more than one cluster can be associated to the same track, the first step is to assimilate all points belonging to one track from the current time step, and recompute the parameters of this larger cluster. The state update is then performed using the parameters of this updated cluster (see Figure~\ref{fig:one-many-association}).

\subsubsection{Track not associated to any cluster}
\label{subsubsec:track-no-cluster}
When an existing track is not associated to any cluster, no updates are performed and the track is flagged to denote that the track had no associations. If the track is flagged for $5$ consecutive time steps, it is assumed to be stale and is deleted from memory.
\section{Velocity Estimation using RLS}
\label{sec:rls}
The instantaneous velocity of a cluster is essential for performing accurate track-cluster association and for assigning a good initial velocity estimate to new tracks so that fast moving vehicles can accurately be tracked by the tracker.

\begin{figure}
  \centering
  \begin{subfigure}[b]{0.45\columnwidth}
  \centering
  	\includegraphics[width=\linewidth]{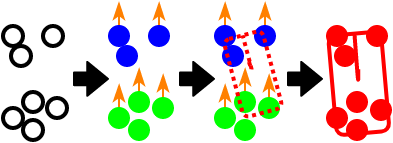}  
    \caption{}
    \label{fig:one-many-association}
  \end{subfigure}
  \hspace{0.5em}
  \begin{subfigure}[b]{0.50\columnwidth}
  \centering
  	\includegraphics[width=\linewidth]{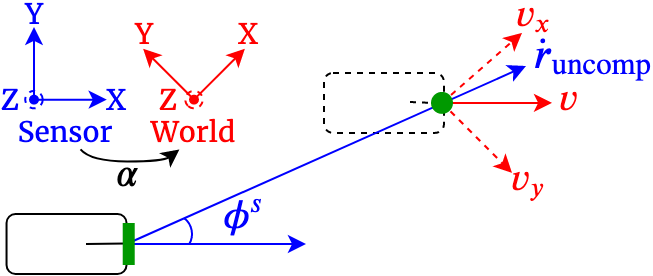}
    \caption{}
    \label{fig:radar-geometry}
  \end{subfigure}
  \setlength{\abovecaptionskip}{-10pt}
  \setlength{\belowcaptionskip}{-20pt}
  \caption{(a) One-many track-cluster association. All points in the image belong to the same object, but DBSCAN initially clusters the top and bottom sets of points into two different clusters. Their computed velocity vectors are shown in orange. When track-cluster association is performed between the clusters and the propagated track (dashed red rectangle), both clusters are associated to the same track resulting in the accurate grouping of points. The heading and extents of the existing track are also corrected in the measurement update and the updated track is shown using a solid red rectangle. (b) Geometry to compute the axis-aligned velocity of the tracked object. Items in the sensor and world frames are depicted using blue and red respectively, and the rotation between them is denoted by $\alpha$. $\phi^{\text{s}}$ is the bearing of the point w.r.t. the radar in the sensor frame, $\dot{r}_{\text{uncomp}}$ is the uncompensated range rate as measured by the radar and does not account for the motion of the ego vehicle, and $v_x$ and $v_y$ are the velocity components of the tracked object in the world frame.}
\end{figure}


\subsection{Problem Formulation}
The velocity of a cluster can be explicitly computed using the Doppler velocity and the bearing w.r.t. the sensor of all points in the cluster. The Doppler velocity defines the rate at which the target is coming towards or moving away from the sensor, and bearing refers to the angle between the ray extending perpendicularly outwards from the sensor and a line pointing directly at the target as shown in Figure~\ref{fig:radar-geometry}. Assuming that all points on the target object have identical velocity vectors, i.e. all objects in the world are rigid, the velocity of a cluster can be computed by solving a linear system of equations as shown in~\eqref{eqn:vel-lin-sys}.
\begin{gather}
\begin{gathered}
\label{eqn:vel-lin-sys}
    \dot{r}_{1} = v_{x}\cos(\phi_{1}^{\text{w}}) + v_{y}\sin(\phi_{1}^{\text{w}}) \\
    \dot{r}_{2} = v_{x}\cos(\phi_{2}^{\text{w}}) + v_{y}\sin(\phi_{2}^{\text{w}}) \\
           \vdotswithin{=} \\
    \dot{r}_{n} = v_{x}\cos(\phi_{n}^{\text{w}}) + v_{y}\sin(\phi_{n}^{\text{w}}) \\
\end{gathered}
\\\phi^{\text{w}} = \phi^{\text{s}} + \theta^{\text{s}} + \alpha
\label{eqn:radar-bearing-world}
\end{gather}
Here, $\dot{r}_{n}$ refers to the compensated range rate of the n\textsuperscript{th} point in the cluster, $v_x$ and $v_y$ correspond to the velocity components of the cluster in the world frame, and $\phi_{n}^{\text{w}}$ is the bearing of the n\textsuperscript{th} cluster point in the world frame computed using~\eqref{eqn:radar-bearing-world}. In~\eqref{eqn:radar-bearing-world}, $\phi^{\text{w}}$ and $\phi^{\text{s}}$ refer to the bearing of the point in the world and sensor frames respectively, $\theta^{\text{s}}$ refers to the angle of the radar w.r.t. the ego-vehicle, and $\alpha$ refers to the rotation angle between the sensor and world frames.

\subsection{Compensated Range Rate}
The compensated range rate, needed for estimating the cluster velocity, can be computed by first computing the range rate of ego-vehicle as experienced by the sensor and then adding the measured range rate as shown in~\eqref{eqn:range-rate-sum}. The range rate of the ego-vehicle is in-turn estimated by first computing the motion at the sensor using the known velocity of the ego-vehicle, and then projecting it onto the line joining the sensor and reflection point as shown in~\eqref{eqn:motion-at-sensor} and~\eqref{eqn:ego-range-rate}.
\begin{gather}
\label{eqn:motion-at-sensor}
    \begin{bmatrix}
    \omega^{\text{s}} \\ v_{x}^{\text{s}} \\ v_{y}^{\text{s}}
    \end{bmatrix}
    = 
    \begin{bmatrix}
    1 & 0 & 0 \\
    -(y^{\text{s}} - y^{\text{e}}) & 1 & 0 \\
    (x^{\text{s}} - x^{\text{e}}) & 0 & 1
    \end{bmatrix}
    \begin{bmatrix}
    \omega^{\text{e}} \\
    v_{x}^{\text{e}} \\
    v_{y}^{\text{e}}
    \end{bmatrix}
    \\
    \label{eqn:ego-range-rate}
    \dot{r}_{\text{e}} = v_{x}^{\text{s}}\cos(\theta^{\text{s}} + \phi^{\text{s}}) + v_{y}^{\text{s}}\sin(\theta^{\text{s}} + \phi^{\text{s}})
    \\
    \label{eqn:range-rate-sum}
    \dot{r}_{\text{comp}} = \dot{r}_{\text{meas}} + \dot{r}_{\text{e}}
\end{gather}
Here $v_{x}^{\text{s}}$, $v_{y}^{\text{s}}$ and $\omega^{\text{s}}$ represent the motion as experienced by the sensor, $x^{\text{s}}$, $y^{\text{s}}$ and $x^{\text{e}}$, $y^{\text{e}}$ denote the position of the sensor and ego-vehicle respectively, $v_{x}^{\text{e}}$, $v_{y}^{\text{e}}$ and $\omega^{\text{e}}$ represent the known motion of the ego vehicle, $\dot{r}_{\text{e}}, \dot{r}_{\text{meas}}, \dot{r}_{\text{comp}}$ denote the ego-vehicle's range rate, measured range rate and compensated range rate respectively, and $\theta^{\text{s}}$ and $\phi^{\text{s}}$ represent the angle of the sensor w.r.t. the car and the bearing of the point w.r.t. the sensor respectively.

\subsection{Solving the System of Linear Equations using RLS}
The system of linear equations shown in~\eqref{eqn:vel-lin-sys} is trivial to solve and a solution can be obtained when there are at least two distinct points in the cluster. The challenge, however, arises from the fact that the radar measurements are very noisy and using solutions from trivial methods results in incorrect velocity estimates. A similar set of linear equations was solved by the authors in~\cite{ref:lat-vel-est} wherein the outliers are rejected using RANSAC and the velocity is computed by using the Levenberg-Marquardt algorithm. This approach, however, is ineffective in our case due to the sparsity of the point cloud, and because the outlier count is comparable to the inlier count in many cases, rendering RANSAC unusable.

\begin{figure}
    \centering
    \includegraphics[width=0.95\columnwidth]{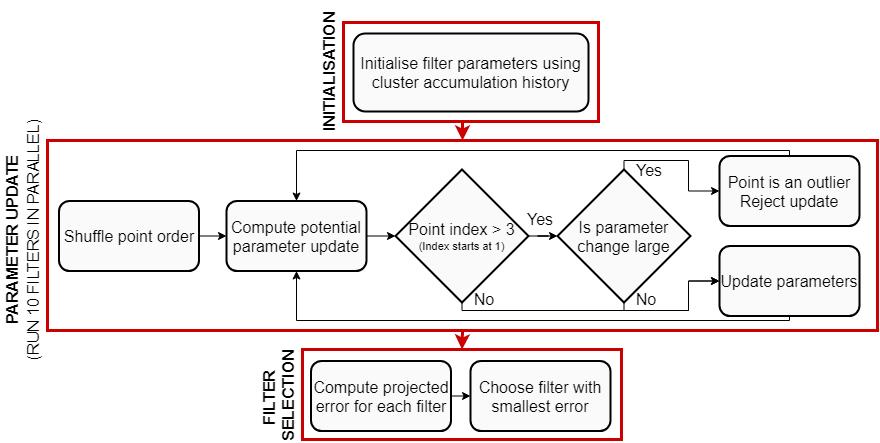}
    \setlength{\abovecaptionskip}{2pt}
    \setlength{\belowcaptionskip}{-20pt}
    \caption{Pipeline of the proposed RLS-based algorithm to compute the instantaneous velocity of a cluster. Image best viewed on a computer.}
    \label{fig:rls-pipeline}
\end{figure}

We instead propose a novel RLS-based approach to compute the velocity of the cluster, the pipeline of which is shown in Figure~\ref{fig:rls-pipeline}. The RLS algorithm is an adaptive filter algorithm that recursively updates the parameters to minimise the weighted least squares cost function. The algorithm starts with an initial estimate of the true velocity which is updated every time a new data point enters the filter. It also exhibits quick convergence and is computationally very fast because it updates the parameters using only the state vector and the new data point. The update equations are shown in~\eqref{eqn:rls-param-update} and~\eqref{eqn:rls-cov-update}.
\begin{align}
    \label{eqn:rls-param-update}
    \bm{v}_{n} &= \bm{v}_{n-1} + \frac{\bm{P}_{n-1}\bm{\phi}_{n}}{1 + \bm{\phi}^T_{n}}(\dot{r}_{n} - \bm{\phi}^T_{n}\bm{v}_{n-1})\\
    \label{eqn:rls-cov-update}
    \bm{P}_{n} &= \bm{P}_{n-1} - \frac{\bm{P}_{n-1} \bm{\phi}_{n} \bm{\phi}^T_{n} \bm{P}_{n-1}}{1 + \bm{\phi}^T_{n} \bm{P}_{n-1} \bm{\phi}_{n}}
\end{align}
Here, $\bm{v}_{n} = \begin{bmatrix} \hat{v}_{x}^{n} & \hat{v}_{y}^{n} \end{bmatrix}$ where $\hat{v}_{x}^{n}$ and $\hat{v}_{y}^{n}$ are the estimated velocity components after using the n\textsuperscript{th} point of the cluster, $\bm{P}$ is a positive-definite matrix that can be thought of as the strength of the parameter update, $\bm{\phi}_{n} = \begin{bmatrix} \cos(\phi_{n}^{w}) & \sin(\phi_{n}^{w})\end{bmatrix}$, where $\phi_{n}^{w}$ is the bearing of the n\textsuperscript{th} point in the world frame, and $\dot{r}_{n}$ is the measured compensated range rate value of that data point.


Literature review suggests that the RLS filter is usually initialised by setting $\bm{v}_{0}$ to $\bm{0}$ and $\bm{P}_{0}$ to some random positive definite matrix, usually an identity matrix multiplied by a small positive constant~\cite{monson-hayes-stat-dsp, simon-haykin-kap, simon-haykin-aft}. It was, however, observed that this initialisation did not converge to the true value in many cases because 1) the number of points was too small, 2) one component of the velocity vector dominated the other, and 3) large outliers in the data pulled the filter away from the true value.
One scheme that works very well is the computation of the initial velocity estimate from the track-history. Since the position estimates from the radar are accurate, the initial velocity can be estimated by computing the change in position and dividing it by the time between the two frames as shown in~\eqref{eqn:rls-init-vel}.
\begin{equation}
    \label{eqn:rls-init-vel}
    \bm{v}_{0}^{t_2} = \frac{\bm{p}^{t_2} - \bm{p}^{t_1}} {t_2 - t_1}
\end{equation}
Here, $\bm{v}_{0}^{t_2}$ refers to the initial velocity estimate at time step $t_2$, $\bm{p}^{t_2}$ and $\bm{p}^{t_1}$ are the positions of the cluster at time steps $t_2$ and $t_1$ respectively. As will be shown in section~\ref{sec:experiments}, this estimate is only good for initialisation and is not accurate enough to be directly used for the track-cluster association and track initialisation.

\subsection{Handling Noise in the Data}
Each cluster usually contains a large number of outlying compensated range rate measurements which can throw off and cause an incorrect convergence of the filter. Such outliers, which are usually a result of reflections from wheel wells and other moving parts of the vehicle, cannot be easily mapped to different parts of a vehicle and thus cannot be removed using trivial geometric methods. We, instead, use the update step of the RLS filter to determine whether a point is an outlier and if its filter update should be used. A point is said to be an outlier if its update results in a change greater than a pre-defined threshold - \SI{0.4}{\meter/\second} in our case - to the $v_x$ and $v_y$ values. This check for outliers is performed only after the first $3$ updates of the filter to allow for some inaccuracies in the initialisation and because it has been noted that most of the filters converge to a value that is close enough to the final estimate in the first $3$ steps. Furthermore, since the order of points entering the filter affects its final output, especially in cases where the first few updates are performed only by outliers which can result in inliers being classified as noise, $10$ filters are initialised for each cluster with each filter being given a randomised order of points as input. To choose the filter with the best parameters, we introduce an error metric called \textit{re-projection error} which computes the sum of errors between the measured and the computed range rates for all points in the cluster marked as inliers. Mathematically,
\begin{gather}
    \label{eqn:comp-range-rate}
    \hat{\dot{r}}_{n} = \hat{v}_{x}\cos(\phi_{n}^{w}) + \hat{v}_{y}\sin(\phi_{n}^{w}) \\
    \label{eqn:reprojection-error}
    e_{n}^{\text{reproj}} = \left|\hat{\dot{r}}_{n} - \dot{r}\right|
\end{gather}
where $\hat{\dot{r}}_{n}$ is the computed range rate for the n\textsuperscript{th} inlier point using the estimated $\hat{v}_{x}$ and $\hat{v}_{y}$ from the filter and $\phi_{n}^{w}$ from the measurement, $|\cdot|$ computes the absolute value, and $e_{n}^{\text{reproj}}$ is the re-projection error for that point. The filter having the least re-projection error amongst the $10$ filters wins and its velocity estimate is chosen to be the velocity of the cluster.

\section{Experiments}
\label{sec:experiments}

\subsection{Experimental Setup}
\label{subsec:experimental-setup}
\begin{figure*}
    \centering
     \includegraphics[width=0.94\textwidth]{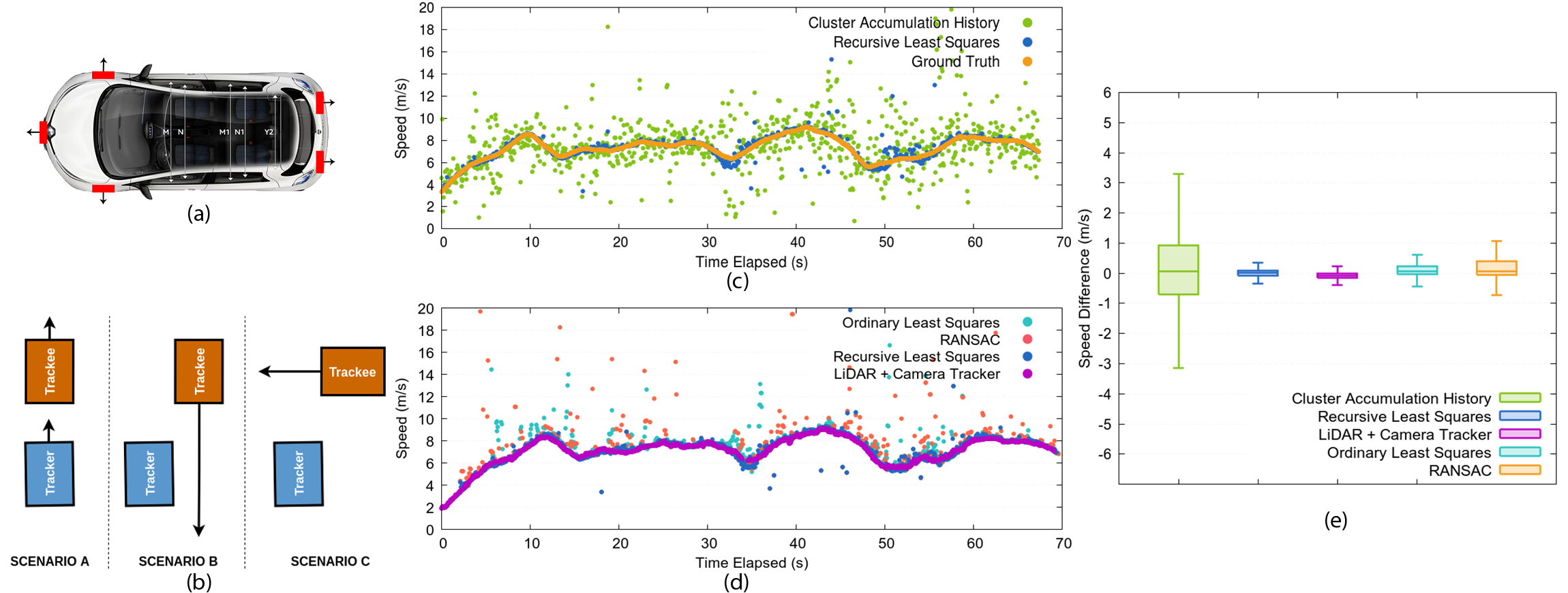}
     \setlength{\abovecaptionskip}{1pt}
     \setlength{\belowcaptionskip}{-20pt}
     \caption{(a) The \SI{77}{\giga\hertz} long-range Continental ARS 408-21 radars in the front and rear bumpers of the Zoe are shown using red rectangles and the arrows denote the perpendicularly outwards direction w.r.t. the sensor. (b) The three scenarios A, B, C used for the experiments. (c, d) qualitatively compare the velocity estimated by our RLS-based algorithm to that of CAH and GT, and that estimated by OLS, RANSAC, and L+C tracker for scenario A respectively. The deviation from GT for each approach is depicted using box-plots in (e). It is evident that our algorithm is consistent with GT, is comparable in performance to the information-rich L+C tracker and notably outperforms other algorithms like OLS and RANSAC followed by OLS.}
     \label{fig:sensor-scenario-a}
\end{figure*}

All the experiments described in this section have been performed using Renault Zoes equipped with multiple \SI{77}{\giga\hertz} long-range Continental ARS 408-21 radars in the front and rear bumpers as shown in Figure~\ref{fig:sensor-scenario-a}a. The radar sensor has a positional resolution of \SI{0.4}{\meter} with an ability to distinguish targets when they are at least \SI{0.6}{\meter} apart, and a velocity resolution of \SI{0.12}{\meter/\s}. Henceforth, the vehicle performing the tracking is referred to as the \textit{tracker}, and the vehicle being tracked is referred to as the \textit{trackee}. The true velocity of the trackee is obtained by using an IMU in conjunction with wheel speed sensors with an accuracy of \SI{0.1}{\meter/\s}, and its true position is obtained by using a LiDAR coupled with pre-annotated HD maps and GPS receivers with centimetre accuracy. The performance of various parts of the algorithm is evaluated using three distinct scenarios as shown in Figure~\ref{fig:sensor-scenario-a}b. In scenario A, the tracker follows the trackee around a $\sim$\SI{750}{\meter} long test track, and in scenarios B and C the tracker is stationary and the trackee drives towards the tracker, and perpendicular to the tracker respectively.


\subsection{Velocity Estimation}
\label{subsec:experiment-velocity}

\begin{table}
\centering
 \begin{tabular}{|c|c|c|c|c|c|c|}
 \hline
 \textbf{Scenario} & & \textbf{CAH} & \textbf{RLS} & \textbf{L+C} & \textbf{OLS} & \textbf{RANSAC}\\
 \hline
 \multirow{3}{*}{\textbf{A}} &
 $\bm{\mu}$ & 0.269 & \color{red}\textbf{0.038}\color{black} & \color{blue}\textbf{-0.102}\color{black} & 0.301 & 0.552  \\ 
 & $\bm{\widetilde{x}}$ & 0.061 & \color{red}\textbf{0.014}\color{black} & -0.071 & \color{blue}\textbf{0.053}\color{black} & 0.054  \\ 
 & $\bm{\sigma^2}$ & 4.950 & \color{blue}\textbf{0.834}\color{black} & \color{red}\textbf{0.044}\color{black} & 1.418 & 4.010  \\
 \hline
 \multirow{3}{*}{\textbf{B}} & 
 $\bm{\mu}$ & 0.458 & \color{blue}\textbf{0.171}\color{black} & \color{red}\textbf{-0.075}\color{black} & 1.461 & 0.239  \\ 
 & $\bm{\widetilde{x}}$ & 0.601 & \color{red}\textbf{-0.013}\color{black} & -0.035 & 0.315 & \color{blue} \textbf{0.025} \color{black}  \\
 & $\bm{\sigma^2}$ & 4.825 & \color{blue}\textbf{0.456}\color{black} & \color{red}\textbf{0.101}\color{black} & 10.108 & 2.369  \\
 \hline
 \multirow{3}{*}{\textbf{C}} & 
 $\bm{\mu}$ & - & \color{blue}\textbf{-0.221}\color{black} & \color{red}\textbf{-0.137}\color{black} & - & -  \\ 
 & $\bm{\widetilde{x}}$ & - & \color{blue}\textbf{-0.137}\color{black} & \color{red}\textbf{-0.131}\color{black} & - & -  \\ 
 & $\bm{\sigma^2}$ & - & \color{blue}\textbf{0.780}\color{black} & \color{red}\textbf{0.088}\color{black} & - & -  \\
 \hline
\end{tabular}
  \setlength{\abovecaptionskip}{2pt}
  \setlength{\belowcaptionskip}{-22pt}
\caption{Mean ($\mu$), median ($\widetilde{x}$), and variance ($\sigma^2$) of the difference in speed  between various algorithms and the ground truth. The smallest and second smallest value in terms of absolute error value are coloured using red and blue respectively. It can be seen that our RLS-based algorithm performs very well and is comparable to the information-rich L+C tracker across all metrics for all scenarios.}
\label{tab:vel-est-stats}
\end{table}

\begin{figure*}
    \centering
     \includegraphics[width=0.94\textwidth]{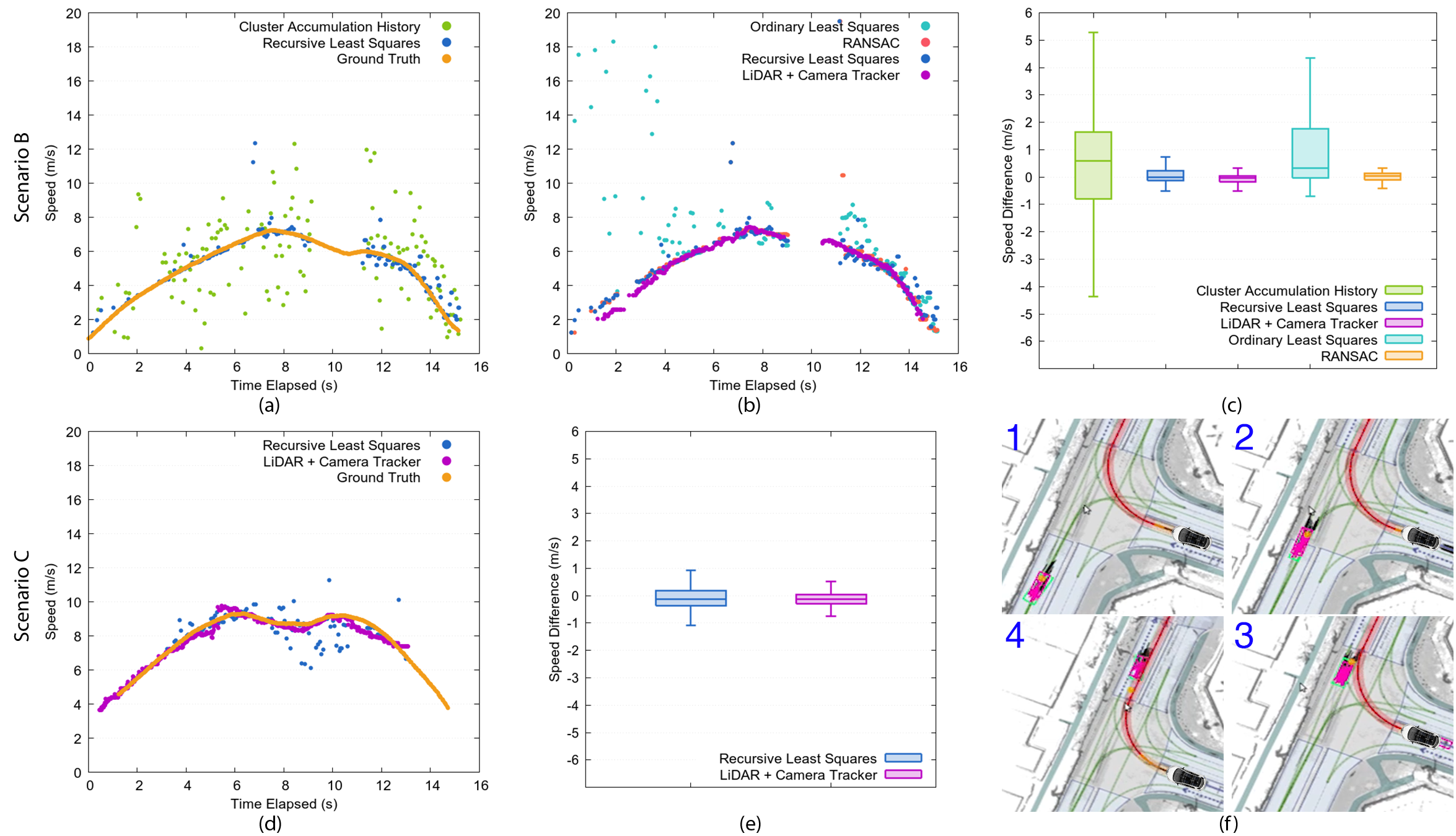}
     \setlength{\abovecaptionskip}{0pt}
     \setlength{\belowcaptionskip}{-18pt}
     \caption{Comparison of the velocity estimated by our RLS-based algorithm to that of (a) the velocity from CAH and GT and (b) the velocity estimated by OLS, RANSAC followed by OLS, and L+C tracker for scenario B. (d) compares the output of our RLS-based algorithm to that from GT and L+C tracker for scenario C. The deviation from GT for each of the approaches in scenarios B and C is depicted using box-plots in (c, e) respectively. Lastly, (f), in the clockwise direction, qualitatively shows the successful tracking of a dynamic object (pink) moving perpendicular to the ego-vehicle using $4$ time-steps of a real-world scenario. Please refer to \href{https://www.youtube.com/watch?v=B5SPXXkZpz8&t=28}{our accompanying video} for more information. Our algorithm agrees with GT in both the scenarios and is significantly better than OLS in Scenario B. RANSAC is able to discard most of the noise points and its performance in scenario B is comparable to that of ours. Lastly, even in the presence of sparse and noisy data, the performance of our algorithm is comparable to that of its information-rich L+C counterpart.}
     \label{fig:scenario-b-c}
\end{figure*}

In this section, we evaluate the performance of our RLS-based velocity estimation algorithm using the scenarios described above. Figures~\ref{fig:sensor-scenario-a}(c-e), \ref{fig:scenario-b-c}(a-c), \ref{fig:scenario-b-c}(d-f) qualitatively compare the performance of our algorithm to the velocity from the cluster accumulation history (CAH) and ground truth (GT), and to widely used algorithms like Ordinary Least Squares (OLS), RANSAC followed by OLS, and also to the velocity estimated from LiDAR + Camera (L+C) fusion for scenarios A, B, and C\footnote{Due to issues with data in scenario C, we compare the performance of our RLS-based algorithm only to the velocities obtained from GT and L+C tracker, both of which were computed online during data collection.} respectively. Table~\ref{tab:vel-est-stats} quantitatively compares the deviation of all algorithms to GT using metrics like mean, median and variance.

It can be inferred from Figure~\ref{fig:sensor-scenario-a}c that the RLS algorithm is able to accurately estimate the velocity of the trackee over extended periods of time. Some deviation can be observed around \SI{32}{\s} when the algorithm underestimates the velocity by around \SI{0.5}{\meter/\s} which can be attributed to an error in the ego-vehicle motion compensation causing the compensated range rate to be lower than the true value. Additionally, some outlier values observed around \SI{52}{\s} are caused by measurements being included from other tracks due the absence of track IDs in the velocity estimation phase. In addition, the extreme values that appear in the plot can be attributed to the few cases wherein the algorithm is unable to reject all noise points and is said to have failed. Lastly, it is worth noting that the velocity obtained from CAH is extremely noisy, but using it as a seed for the RLS filter results in very accurate velocity estimates. Furthermore, it is evident from Figure~\ref{fig:sensor-scenario-a}d that our algorithm outperforms both OLS and RANSAC followed by OLS, which can be verified from Figure~\ref{fig:sensor-scenario-a}e and Table~\ref{tab:vel-est-stats} wherein our algorithm has a smaller deviation from GT, and also a lower inter-quantile-distance and variance as compared to OLS and RANSAC followed by OLS. These approaches fail because of the presence of large outliers that are not effectively handled by either of them due to the inherent lack of outlier rejection, sparsity of the point cloud, and a large outlier count. It is also noted that RANSAC performs worse than OLS due to a large outlier count causing RANSAC to choose noise points as inliers which results in multiple incorrect velocity estimates. OLS suffers to a lesser extent because the true inliers pull the velocity estimate closer to GT resulting in lower error metrics. Figures~\ref{fig:sensor-scenario-a}d, \ref{fig:sensor-scenario-a}e and Table~\ref{tab:vel-est-stats} show that the velocity computed by our algorithm is comparable to that estimated by the L+C tracker which works by first associating objects over frames using both LiDAR and cameras and then computing the velocity using an Extended Kalman Filter on the dense LiDAR point cloud. The variance, however, is much larger for our algorithm due to the few instances when all the noise points are not effectively filtered out resulting in the RLS filter computing very large and incorrect velocity estimates.


It can be inferred from Figures~\ref{fig:scenario-b-c}a and \ref{fig:scenario-b-c}d that our algorithm is able to accurately estimate the velocity of the trackee during both the acceleration and deceleration phases. The gap in the output from \SI{9}{\s} to \SI{11}{\s} in Figure~\ref{fig:scenario-b-c}a is due to the trackee being closer to the tracker than the minimum allowed distance resulting in no measurements from any sensor. The extreme values, like the two points at \SI{7}{\s}, are due to the filter not rejecting all the outliers and can be classified as a failure of our algorithm. Furthermore, the few erroneous velocity estimates between \SI{9}{\s} and \SI{11}{\s} in Figure~\ref{fig:scenario-b-c}d are due to both the inclusion of the velocity estimates from other clusters, and the trackee being exactly perpendicular to the tracker resulting in valid range rate estimates being filtered out by our pre-processing step due to them being very close to \SI{0}{\meter/\s}.
Lastly, the superior performance of our algorithm is also evident in Figures~\ref{fig:scenario-b-c}b and \ref{fig:scenario-b-c}c wherein OLS fails due to the lack of noise handling capabilities. RANSAC, on the other hand, is able to reject most of the outliers resulting in its performance being comparable to that of ours in scenario B. Furthermore, it can be observed that in scenarios B and C, the velocity from the L+C tracker is better than that estimated by our algorithm because of the improved position estimates between frames due to the presence of a larger ``width" view of objects in the point cloud. The radar, however, does not benefit from this because the velocity is computed directly using the measured Doppler velocity of the points on the object. Nevertheless, as seen in Table~\ref{tab:vel-est-stats}, the performance of our RLS-based algorithm is still very accurate and is numerically comparable to the fusion approach. 


\subsection{Tracking Performance}
\label{subsec:experiment-tracking-performance}
\begin{figure}[h]
	\centering
  \begin{subfigure}{\columnwidth}
  	\centering
   	\includegraphics[width=0.8\linewidth]{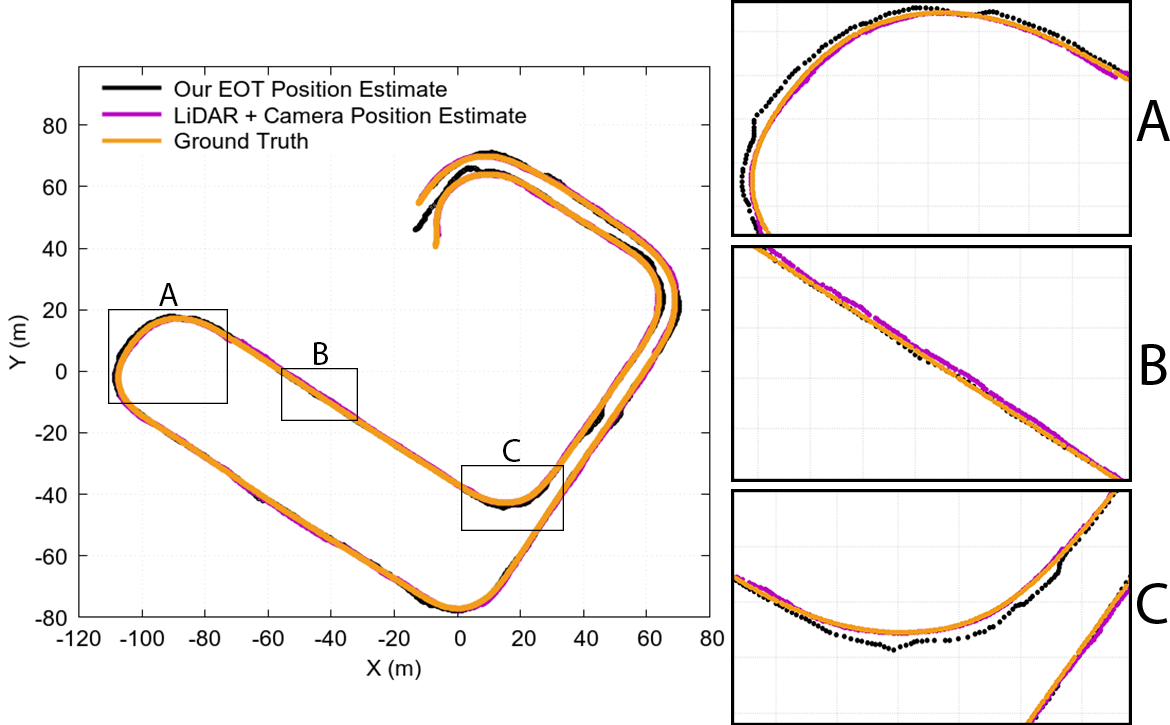}
    \setlength{\abovecaptionskip}{0pt}
    \setlength{\belowcaptionskip}{-2pt}
    \caption{}
    \label{fig:loop-gt-trk-comp}
  \end{subfigure}
  \begin{subfigure}[b]{\columnwidth}
    \centering
   	\includegraphics[width=0.8\linewidth]{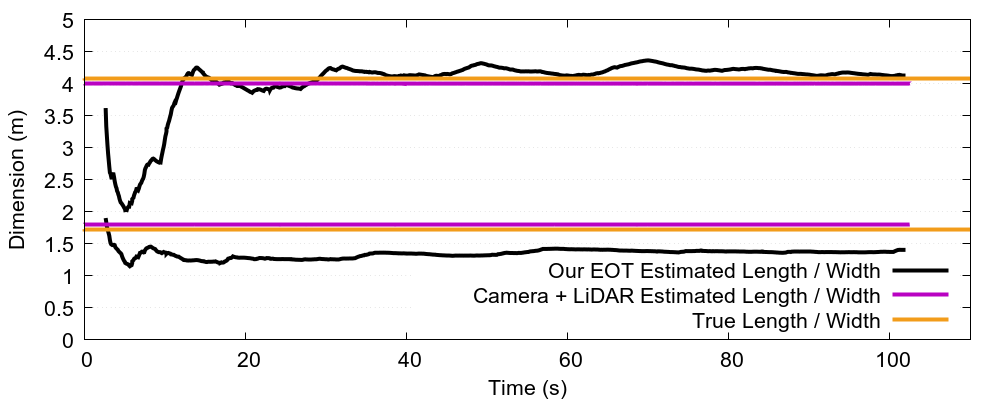}  
    \setlength{\abovecaptionskip}{0pt}
    \setlength{\belowcaptionskip}{-2pt}
    \caption{}
    \label{fig:loop-gt-extents-comp}
  \end{subfigure}
  \setlength{\abovecaptionskip}{-8pt}
  \setlength{\belowcaptionskip}{-24pt}
  \caption{Tracking is evaluated using two metrics, namely, position and extents. (a) overlays GT of trackee over the tracked estimates from our EOT. Very few deviations from GT are observed proving the positional accuracy of our tracker. (b) compares the estimated extents with the ground truth, from which we infer that although the length is quite accurate, the width is always underestimated due to the dearth of points in the width-wise direction.}
\end{figure}

This section evaluates the tracking performance of our EOT using two metrics, i.e., positional tracking accuracy, and extents tracking accuracy, using scenario A. Figure~\ref{fig:loop-gt-trk-comp} compares the trackee's position computed by the tracker to that of GT. It is noted that the tracker is very accurate along straights but overestimates the radius during turns. This deviation can be attributed to the high measurement noise we use to account for the inaccuracies in the radar which results in the update step favouring the propagated state over the new measurements. Because the velocity vector is tangential to the curve, the prediction step pushes the predicted position out of the circle which is only partially corrected by the measurement update resulting in an overestimation of the radius of the curve. Such an overestimation can be addressed by using more sophisticated models like Constant Turn Rate Velocity (CTRV) and so on.
Furthermore, the deviations on straights like those observed at $(45, -15)$ and $(-65, -45)$ are due to several erroneous points appearing on the road surface. Because of the proximity of these points to true ones, they are clustered together resulting in an incorrect measurement update and subsequently an incorrect track position. The L+C tracker, in contrast, tracks the trackee very precisely and its prediction almost overlaps with that of GT, which can be attributed to the presence of a very dense and noise-free 3D point cloud from the LiDAR.

Figure~\ref{fig:loop-gt-extents-comp} compares the extents estimate of the trackee to GT. As previously stated, the length and width of the track are initialised to \SI{4}{\meter} and \SI{2}{\meter} respectively when the track is created and are updated as part of the measurement update step. The initial sharp drop can be explained by the lack of measurements with range rate greater than \SI{0.5}{\meter/\s} resulting in the extents being severely underestimated. Once the cars start moving more point pass through the filter which immediately improves the dimension estimates. Lastly, due to the large angular resolution of the radar which limits the number of returns from an object width-wise, the width is always underestimated by our tracker. This can be solved by either adding a bias, using radars with a smaller angular resolution, or using a model that accounts for such a behaviour.
The L+C tracker predicts the extents of the car very accurately due to the presence of an information-rich and noise-free point cloud. The extents are constant throughout because the algorithm does not update the extents once a ``good-enough'' view of the trackee is obtained.

\begin{figure}
	\centering
   	\includegraphics[width=0.9\linewidth]{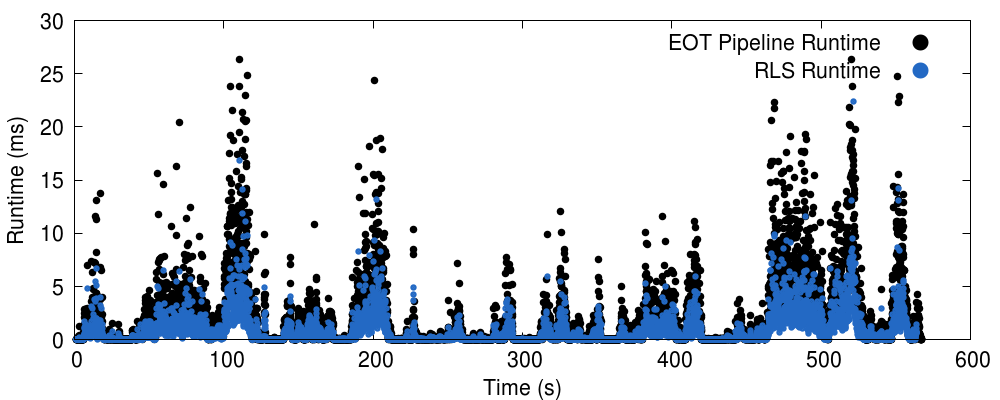}
  \setlength{\abovecaptionskip}{0pt}
  \setlength{\belowcaptionskip}{-18pt}
  \caption{Runtime of both the EOT pipeline and the RLS-based velocity estimation module. It can be noted that both of them run in real-time, thus allowing for their direct implementation on autonomous vehicles.}
  \label{fig:one-north-runtime}
\end{figure}

\subsection{Runtime}
\label{subsec:experiment-runtime}

Figure~\ref{fig:one-north-runtime} plots the runtime of both the extended tracking pipeline, and the RLS-based velocity estimation module in real-world traffic with the number of objects ranging from $5$ to $30$ per frame. It is noted that the RLS module runs in real-time averaging \SI{0.89}{\milli\s} for all clusters in a frame. Similarly, the entire pipeline also runs in real-time averaging \SI{2.1}{\milli\s} per frame with all frames being computed in less than \SI{30}{\milli\s}. We can thus safely conclude that our EOT pipeline can run in real-time even in dense traffic scenarios.

\section{Conclusion}
\label{sec:conclusion}
In this paper we presented a novel RLS-based approach to estimate the instantaneous velocity of a vehicle in real-time. The RLS filter uses the compensated range rate and the bearing w.r.t. the sensor of all points in the cluster to solve a linear system of equations to estimate the velocity. Because the range rate estimates are very noisy, an efficient outlier rejection module was built into the filter to separate the inliers from the outliers. It was shown that our approach performs significantly better than the other widely used approaches such as OLS and RANSAC followed by OLS, and is comparable to the performance of the information-rich LiDAR + Camera tracker. Furthermore, we also presented a pipeline to track extended objects in real-time which uses the computed velocity estimates to initialise new tracks and perform track-cluster association. This tracker runs in real-time and was also shown to be very precise enabling its direct implementation on autonomous vehicles. Some shortcomings of our approach like the overestimation of radius during turns and the underestimation of the width of objects can be resolved either by using more sophisticated models that account for these intricacies, or by using a higher resolution and more accurate radar sensor, and we intend to address them in a future work.


%

\newrefcontext[sorting=none]
\printbibliography

@INPROCEEDINGS{ref:lat-vel-est, 
author={D. {Kellner} and M. {Barjenbruch} and K. {Dietmayer} and J. {Klappstein} and J. {Dickmann}}, 
booktitle={Proceedings of the 16th International Conference on Information Fusion}, 
title={Instantaneous lateral velocity estimation of a vehicle using Doppler radar}, 
year={2013}, 
volume={}, 
number={}, 
pages={877-884}, 
doi={}, 
ISSN={}, 
month={July},}

@inproceedings{ref:dbscan,
  title={A density-based algorithm for discovering clusters in large spatial databases with noise.},
  author={Ester, Martin and Kriegel, Hans-Peter and Sander, J{\"o}rg and Xu, Xiaowei and others}
}

@misc{Scheel.2018,
author = {Scheel, Alexander and Dietmayer, Klaus},
year = {2018},
title = {{Tracking Multiple Vehicles Using a Variational Radar Model}},
url = {https://arxiv.org/abs/1711.03799},
archivePrefix = {arXiv},
primaryClass = {eess.SP},
eprint = {1711.03799v2}
}

@article{eotsummary,
author = {Granstrom, Karl and Baum, Marcus and Reuter, Stephan},
year = {2017},
month = {12},
pages = {},
title = {Extended Object Tracking: Introduction, Overview, and Applications},
volume = {12},
journal = {Journal of Advances in Information Fusion}
}

@inproceedings{Henriksson2016RadarBT,
  title={Radar based target tracking for 360-degree environmental perception},
  author={Erik Henriksson and Viktor Kardell},
  year={2016}
}

@ARTICLE{6237597, 
author={L. {Hammarstrand} and L. {Svensson} and F. {Sandblom} and J. {Sorstedt}}, 
journal={IEEE Transactions on Aerospace and Electronic Systems}, 
title={Extended Object Tracking using a Radar Resolution Model}, 
year={2012}, 
volume={48}, 
number={3}, 
pages={2371-2386}, 
doi={10.1109/TAES.2012.6237597}, 
ISSN={0018-9251}, 
month={JULY},}

@article{Scheel2016MultisensorMT,
  title={Multi-sensor multi-object tracking of vehicles using high-resolution radars},
  author={Alexander Scheel and Christina Knill and Stephan Reuter and Klaus C. J. Dietmayer},
  journal={2016 IEEE Intelligent Vehicles Symposium (IV)},
  year={2016},
  pages={558-565}
}

@inproceedings{dirscattering,
author = {Knill, Christina and Scheel, Alexander and Dietmayer, Klaus},
year = {2016},
month = {06},
pages = {298-303},
title = {A direct scattering model for tracking vehicles with high-resolution radars},
doi = {10.1109/IVS.2016.7535401}
}

@ARTICLE{7161279,
author={J. {Dickmann} and N. {Appenrodt} and J. {Klappstein} and H. {Bloecher} and M. {Muntzinger} and A. {Sailer} and M. {Hahn} and C. {Brenk}},
journal={IEEE Access},
title={Making Bertha See Even More: Radar Contribution},
year={2015},
volume={3},
number={},
pages={1233-1247},
doi={10.1109/ACCESS.2015.2454533},
ISSN={2169-3536},
month={},}

@INPROCEEDINGS{7535453,
author={J. {Elfring} and R. {Appeldoorn} and M. {Kwakkernaat}},
booktitle={2016 IEEE Intelligent Vehicles Symposium (IV)},
title={Multisensor simultaneous vehicle tracking and shape estimation},
year={2016},
volume={},
number={},
pages={630-635},
doi={10.1109/IVS.2016.7535453},
ISSN={},
month={June},}

@ARTICLE{7518649,
author={F. {Roos} and D. {Kellner} and J. {Dickmann} and C. {Waldschmidt}},
journal={IEEE Transactions on Microwave Theory and Techniques},
title={Reliable Orientation Estimation of Vehicles in High-Resolution Radar Images},
year={2016},
volume={64},
number={9},
pages={2986-2993},
doi={10.1109/TMTT.2016.2586476},
ISSN={0018-9480},
month={Sep.},}

@INPROCEEDINGS{7918865,
author={J. {Schlichenmaier} and N. {Selvaraj} and M. {Stolz} and C. {Waldschmidt}},
booktitle={2017 IEEE MTT-S International Conference on Microwaves for Intelligent Mobility (ICMIM)},
title={Template matching for radar-based orientation and position estimation in automotive scenarios},
year={2017},
volume={},
number={},
pages={95-98},
doi={10.1109/ICMIM.2017.7918865},
ISSN={},
month={March},}

@ARTICLE{1512732,
author={K. {Gilholm} and D. {Salmond}},
journal={IEE Proceedings - Radar, Sonar and Navigation},
title={Spatial distribution model for tracking extended objects},
year={2005},
volume={152},
number={5},
pages={364-371},
doi={10.1049/ip-rsn:20045114},
ISSN={1350-2395},
month={Oct},}

@INPROCEEDINGS{6641184,
author={K. {Granstrom} and C. {Lundquist}},
booktitle={Proceedings of the 16th International Conference on Information Fusion},
title={On the use of multiple measurement models for extended target tracking},
year={2013},
volume={},
number={},
pages={1534-1541},
doi={},
ISSN={},
month={July},}

@ARTICLE{6237574,
author={L. {Hammarstrand} and M. {Lundgren} and L. {Svensson}},
journal={IEEE Transactions on Aerospace and Electronic Systems},
title={Adaptive Radar Sensor Model for Tracking Structured Extended Objects},
year={2012},
volume={48},
number={3},
pages={1975-1995},
doi={10.1109/TAES.2012.6237574},
ISSN={0018-9251},
month={JULY},}

@INPROCEEDINGS{4290130,
author={J. {Gunnarsson} and L. {Svensson} and L. {Danielsson} and F. {Bengtsson}},
booktitle={2007 IEEE Intelligent Vehicles Symposium},
title={Tracking vehicles using radar detections},
year={2007},
volume={},
number={},
pages={296-302},
doi={10.1109/IVS.2007.4290130},
ISSN={1931-0587},
month={June},}

@book{monson-hayes-stat-dsp,
author = {Hayes, Monson H.},
title = {Statistical Digital Signal Processing and Modeling},
chapter = {9.4},
pages = {541-546},
year = {1996},
isbn = {0471594318},
publisher = {John Wiley Sons, Inc.},
address = {USA},
edition = {1st}
}

@book{simon-haykin-kap,
author = {Liu, Weifeng and Principe, Jose C. and Haykin, Simon},
title = {Kernel Adaptive Filtering: A Comprehensive Introduction},
chapter = {4.1},
pages = {94-96},
year = {2010},
isbn = {0470447532},
publisher = {Wiley Publishing},
edition = {1st}
}

@book{simon-haykin-aft,
author = {Haykin, Simon},
title = {Adaptive Filter Theory (3rd Ed.)},
chapter = {13},
pages = {569-570},
year = {1996},
isbn = {013322760X},
publisher = {Prentice-Hall, Inc.},
address = {USA}
}

@misc{nvidia-radar,
  title = {Autonomous Vehicle Radar Perception in 360 Degrees},
  howpublished = {\url{https://devblogs.nvidia.com/autonomous-vehicle-radar-perception-in-360-degrees/}},
  note = {Accessed: 2020-02-01}
}

@INPROCEEDINGS{dickmann-radar,
author={J. {Dickmann} and J. {Klappstein} and M. {Hahn} and N. {Appenrodt} and H. {Bloecher} and K. {Werber} and A. {Sailer}},
booktitle={2016 IEEE Radar Conference (RadarConf)},
title={"Automotive radar the key technology for autonomous driving: From detection and ranging to environmental understanding"},
year={2016},
pages={1-6},
ISSN={2375-5318},
month={May},}

@INPROCEEDINGS{wei-radar,
author={J. Wei and J. M. Snider and J. Kim and J. M. Dolan and R. Rajkumar and B. Litkouhi},
booktitle={2013 IEEE Intelligent Vehicles Symposium (IV)},
title={Towards a viable autonomous driving research platform},
year={2013},
pages={763-770},
ISSN={1931-0587},
month={June},
}

@misc{yurtsever-radar,
    title={A Survey of Autonomous Driving: Common Practices and Emerging Technologies},
    author={Ekim Yurtsever and Jacob Lambert and Alexander Carballo and Kazuya Takeda},
    year={2019},
    archivePrefix={arXiv},
    primaryClass={cs.RO},
}

@article{ransac,
author = {Fischler, Martin A. and Bolles, Robert C.},
title = {Random Sample Consensus: A Paradigm for Model Fitting with Applications to Image Analysis and Automated Cartography},
year = {1981},
issue_date = {June 1981},
publisher = {Association for Computing Machinery},
address = {New York, NY, USA},
volume = {24},
number = {6},
issn = {0001-0782},
url = {https://doi.org/10.1145/358669.358692},
doi = {10.1145/358669.358692},
journal = {Commun. ACM},
month = jun,
pages = {381–395},
numpages = {15}
}
\end{document}